# Language-Dependent Political Bias in AI: A Study of ChatGPT and Gemini

Dogus Yuksel[1], Mehmet Cem Catalbas[2], Bora Oc[3]

[1,3]Department of Marketing and Advertising, Ostim Technical University, 06374 Ankara, Turkey

[2]Department of Software Engineering, Ankara University, 06100 Ankara, Turkey

**Abstract:** As leading examples of large language models, ChatGPT and Gemini claim to provide accurate and unbiased information, emphasizing their commitment to political neutrality and avoidance of personal bias. This research investigates the political tendency of large language models and the existence of differentiation according to the query language. For this purpose, ChatGPT and Gemini were subjected to a political axis test using 14 different languages. The findings of the study suggest that these large language models do exhibit political tendencies, with both models demonstrating liberal and leftist biases. A comparative analysis revealed that Gemini exhibited a more pronounced liberal and left-wing tendency compared to ChatGPT. The study also found that these political biases varied depending on the language used for inquiry. The study delves into the factors that constitute political tendencies and linguistic differentiation, exploring differences in the sources and scope of educational data, structural and grammatical features of languages, cultural and political contexts, and the model's response to linguistic features. From this standpoint, and an ethical perspective, it is proposed that artificial intelligence tools should refrain from asserting a lack of political tendencies and neutrality, instead striving for political neutrality and executing user queries by incorporating these tendencies.

## 1. Introduction

In recent years, advances in artificial intelligence (AI) have revolutionized many areas of human life. In particular, Large Language Models (LLMs) have become powerful tools for text generation, analysis and knowledge-based solutions thanks to their natural language processing capabilities. However, the proliferation of these technologies has also raised important questions about their ethical and social implications. The potential political biases inherent in

LLMs can affect both the user experience and the capacity of these models to provide unbiased information. In particular, how political tendencies are reflected in the responses of these models has raised the question of how these technologies can shape social and political structures.

In this context, it is of great importance to understand the political biases of LLMs and to investigate whether these biases change depending on linguistic queries. The political bias of AI-based tools may risk misleading users or contributing to social polarization. In particular, the question of how these tendencies change in different languages is a critical research area in terms of revealing the impact of linguistic and cultural contexts on artificial intelligence models.

With the introduction of artificial intelligence tools for individual use, various studies have been conducted on the political orientation and axes of these tools. For example, following the launch of ChatGPT, their political orientation has been investigated through different tests, such as political surveys on German and Dutch politics (van den Broek, 2023; Hartmann, et al., 2023). According to the discovered results, ChatGPT prefers left-wing parties with a focus on social democrats and environmentalists. Furthermore, it was noted that the answers of ChatGPT differed in terms of attitude towards different groups depending on their political orientation (McGee, 2023a; McGee, 2023b; Rozado, 2023a). Comparing ChatGPT's progressive bias with 15 political affiliation tests, it was concluded that progressive bias was identified in 14 tests (Rozado, 2023b). In research by Rutinowski et al. (2024), OpenAI's Large Language Model examined its self-perception and political biases to assess ChatGPT's claims of predisposition to progressive and libertarian viewpoints. The findings suggest that these biases may have important implications for AI ethics and societal impact and warrant further research. In another study, researchers took ChatGPT through a political compass test along with a US Democrat, a US Republican, and other participants. The Republican and Democratic respondents' profiles were quite predictable within their political affiliations; however, ChatGPT chose responses that were mainly Democratic (Motoki, et al., 2024). In addition, the research conducted by Choudhary (2024) compares the political biases of the Pew Research Center's Political Typology Test, Political Compass and ISideWith test and the political biases of the artificial intelligence model, which covers that ChatGPT-4 and Claude exhibit liberal, Perplexity conservative and Gemini (Google Bard) centrist tendencies.

This study aims to examine the political leanings of LLM platforms and whether these leanings differ across platforms. It also investigates how LLMs' responses to queries in different languages affect their political leanings and provides a comprehensive analysis of their capacity to provide impartial and fair information.

## 2. Artificial Intelligence

Artificial intelligence (AI) is a field of technology that enables systems to do important tasks of varying complexity by mimicking the outcomes of human intelligence (Mondal, 2020). The 1950s marked the development of artificial intelligence as a core notion that is now used in many other domains (Lin and Yu, 2024). The concept of "Artificial Intelligence" was first officially used at the Dartmouth Conference in 1956 and aroused a lot of interest (Cordeschi, 2007). Afterward, research in this field gained significant momentum. When Joseph Weizenbaum created the computer software ELIZA in 1966 and it gained attention for its ability to speak, a significant advancement was accomplished in the science of artificial intelligence (Shum, et al., 2018). Thanks to its natural language processing capabilities, ELIZA had very successful competencies for its time in entering into dialog with humans. ELIZA analyzes the structure of user-generated input text and uses predefined matching techniques to identify specific keywords. Based on these keywords, it selects predefined answers and reports back to the user. The ELIZA computer software can be described as an early application of artificial intelligence, an important forerunner of current LLM frameworks (Chang, et al., 2024). Artificial intelligence finds widespread application in video and audio processing, natural language processing, robotics, autonomous systems and medical technology (Anantrasirichai and Bull, 2022). Depending on the complexity of the problems, AI models are trained on data sets of different sizes. These models can obtain the ability to perform specific tasks by deriving meaningful relationships from the relevant data. Innovations in AI have led to significant changes and advances in sectors such as healthcare, financial services, educational institutions, logistics and many more (Goodell, et al., 2021; Woschank, et al., 2020). For example, various artificial intelligence models are used to improve diagnostic performance in medical fields, to make sense of the data obtained within the scope of different sensor technology applications, or to improve the holistic performance of systems (Briganti and Le Moine, 2020; Catalbas and Dobrisek, 2023). Since the concept of artificial intelligence was introduced, various sub-fields have been formed in a hierarchical relationship depending on the developments in the related field. In this development process, the concept of machine learning as a hierarchical subset comes first.

The concept of machine learning describes the ability of a model to learn relationships from data sets and apply this learning to new data to make new and meaningful predictions. With machine learning, meaningful and effective relationships between input data are learned, and then various decisions and applications are made using these relationships (Kühl, et al., 2022). In the process of obtaining the relevant pattern, a large amount of data and computational power is generally needed. Depending on the meaningful input data and computational power of the system, the performance of the machine learning model increases. Machine learning is performed with different approaches. The most important of these is the training of models using labeled data, which is defined as supervised learning (Alloghani, et al., 2020). Another approach is unsupervised learning, in which the meaningful pattern in the input data is identified and parsed with unlabeled data. Another learning approach is reinforcement learning, in which an agent interacts with the environment and learns the environment with a reward and punishment approach and creates a machine learning model that performs meaningful behaviors (Szepesvári, 2022). The reinforcement learning approach is mostly preferred in systems with high interaction with the physical world such as robotics and automation (Singh, et al., 2022).

Deep learning, which is a subset of machine learning, unlike machine learning, uses multi-layered artificial neural network layers similar to the brain structure to effectively learn complex and high-level features among input data (Dong, et al., 2021). Deep neural networks optimize the weights of the artificial neural network model using input data through feedforward and backpropagation algorithms and increase the performance of the model by using various activation functions (Taye, 2023). Deep learning processes generally use large data sets and powerful computational resources. It provides very successful results in areas such as image and audio processing and natural language processing (Li, et al., 2021). Despite the high computational power requirement compared to traditional machine learning models, it provides considerably high performance. Another subset of AI is termed "generative AI," and it refers to sophisticated AI models that can create new and creative content like text, video, and music while using relationships they have already learned (Feuerriegel, et al., 2024). Deep learning techniques are usually used to create these models. The core of generative artificial intelligence is comprised of neural network architectures, such as variational autoencoders (VAEs) and generative adversarial networks (GANs) (Singh and Ogunfunmi, 2021; Alqahtani, et al., 2021). Producing original content for a variety of media, such as games, movies, videos, news, and art, is made easy and effective by generative artificial intelligence. Prominent instances of generative AI are Large Language Model (LLM) frameworks, which are equipped with the ability to generate and develop languages (Yao, et al., 2024).

Hierarchical Relationship of Artificial Intelligence and Its Sub-Fields is given in Figure 1.

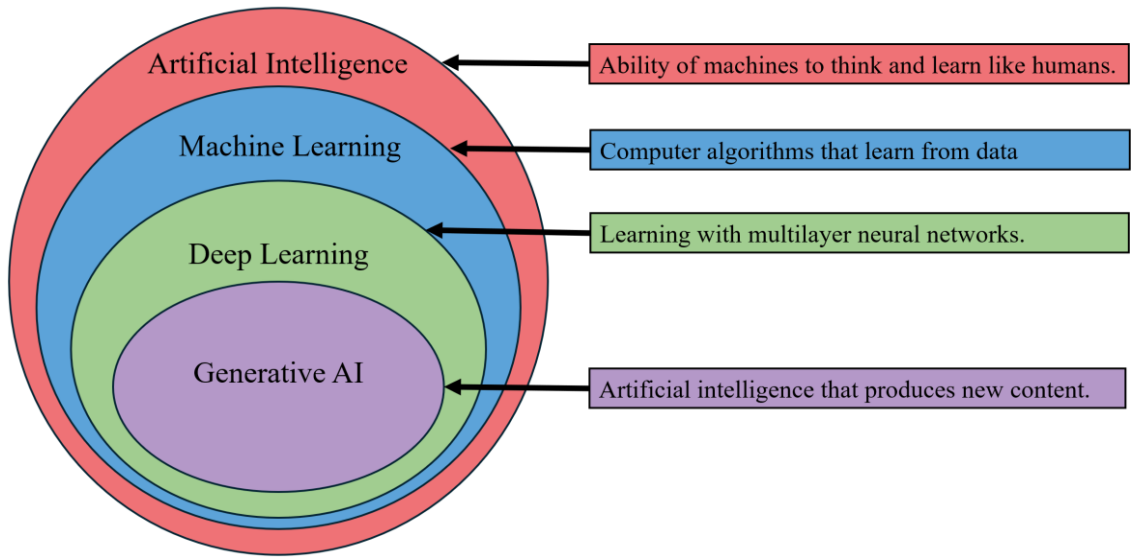

**Figure 1.** Layers of Artificial Intelligence

**2.1. Large Language Model**

Large Language Models (LLMs) are a class of specialized artificial intelligence models designed to allow huge volumes of textual information to be analyzed and processed to understand human language and then to produce solutions (Chang, et al., 2024). Deep learning techniques with specific transformer deep neural network topologies are used by LLMs to achieve impressive accuracy and efficiency in a range of language-based applications. LLMs use books, web pages, scientific articles, news articles, social media posts and extensive web search data as data sources. LLM training utilizes supervised and unsupervised learning, reinforcement learning and transfer learning approaches. LLM architectures include transform models, recurrent neural networks (RNNs), long short-term memory networks (LSTMs) and attention mechanisms (Ignacio, et al., 2024; Yang, et al., 2022). The computation of LLM models utilizes highly computationally capable hardware such as GPUs and TPUs, cloud services and distributed computing structures so that large amounts of input data can be processed and interpreted in reasonable time (Raiaan, et al., 2024). Human feedback and adversarial testing are the preferred approaches for evaluating the performance of the outputs, fine-tuning the model and evaluating the performance and efficiency of LLMs. The generic structure of LLMs with subcomponents is given in Figure 2.

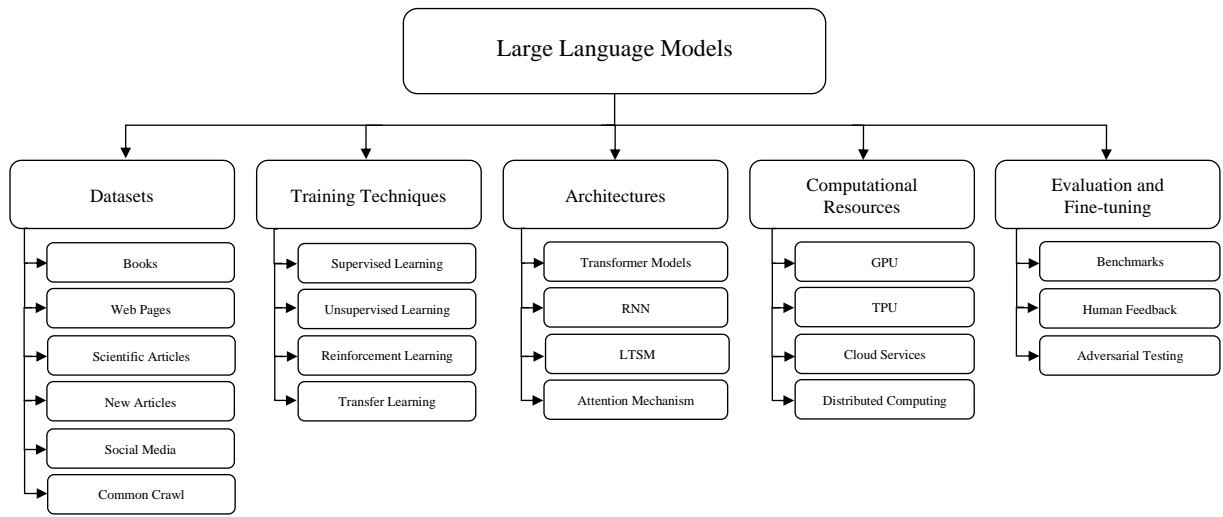

**Figure 2.** Generic Framework of Large Language Models

ChatGPT 3.5 is an advanced LLM developed by OpenAI, capable of generating fluent and human-like text (Ngo, et al., 2024). It has been trained on a large dataset of around 45 terabytes, including academic articles, books, websites, etc., and the timeliness of the training data is limited to September 2021 (Roumeliotis and Tselikas, 2023). However, it is a free model that supports interactions up to 4096 tokens and stands out among LLM models (Kondurkar, et al., 2024). Another LLM model used in this study is Gemini 1.0 Pro developed by Google, which has the ability to create human-like text and code and respond to specific reviews (Carlà, et al., 2024). Unlike ChatGPT, Gemini LLM has the ability to access real-time data instead of a static database. Among these, it has the ability to access current websites and Google applications, and thus has the ability to dynamically scan and update databases. The Gemini language model uses a machine learning model called Mixture of Experts (MoE) (Xie, et al., 2023). The relevant model separates each data input into homogeneous regions and transmits them to separate subnetworks that specialize in certain features, thus effectively learning the data.

### 2.2. Political Compass

The political compass is a versatile tool used to assess individuals' political preferences, offering a comprehensive overview of their attitudes and beliefs on a wide range of political and social issues (Laméris, et al., 2018). The terms "left" and "right," first adopted by the French Estates General in 1789, have been central to political discourse, evolving in meaning over time (Lester, 1994). In order to understand the political orientation of individuals, Leonard W. Ferguson (1941) and Hans Eysenck (1957) first created models for factor analysis of political beliefs in the 1940s and made the first attempts to measure political opinion.

Further complexity was introduced by Bryson and McDill (1968) (See Figure 3), who proposed a bi-dimensional model that added a statist-anarchist axis to the traditional left-right spectrum (Pedraza, et al., 2021). This approach allows for a more sophisticated analysis of political ideologies by examining both the degree of individual freedom and the extent of government control (Bryson and McDill, 1968). Similarly, Christie and Meltzer (1970) expanded this framework by evaluating political systems along axes ranging from authoritarianism to anarchism and from capitalism to socialism, further enriching the discourse on political orientation.

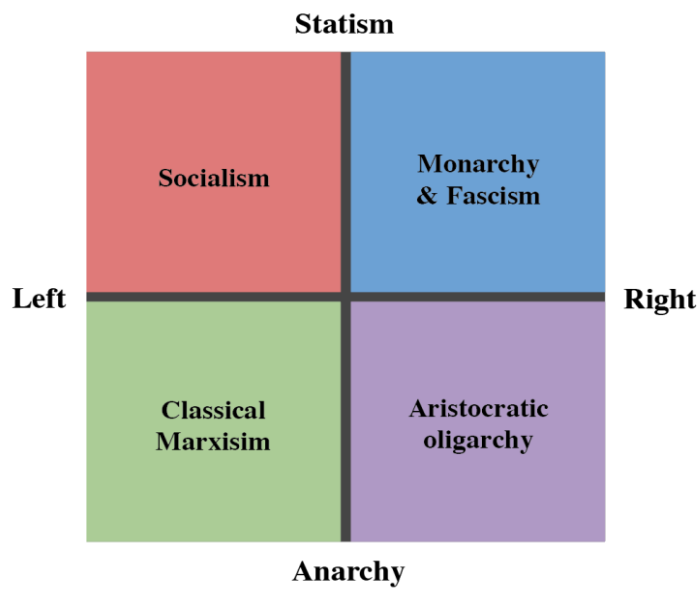

**Figure 3:** Bryson and McDill's A Bi-dimensional Model (1968)

David F. Nolan's (1971) introduction of a two-dimensional chart in 1969 marked a significant development in political analysis by contrasting economic and personal freedoms (Elkind, et al., 2017; Gołębiowska and Sznajd-Weron, 2021), offering a deeper understanding of political beliefs. The Nolan Graph depicted in Figure 4 designates "economic freedom" as the horizontal axis and "personal freedom" as the vertical axis. Five zones make up the graph's square form, and each zone stands for a certain political viewpoint. Nolan's model, known as the Nolan Chart, highlights five distinct zones, each representing different political ideologies based on varying levels of economic and personal freedom (Nolan, 1971).

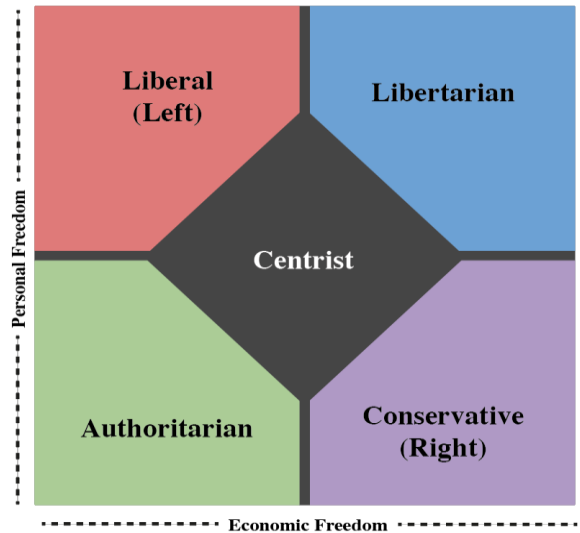

**Figure 4:** The Nolan Chart in Its Traditional Form

Source: *The Nolan Chart. (Seal, 2013)*

In their book "The Floodgates of Anarchy," Stuart Christie and Albert Meltzer present a two-pronged method for comprehending the political divide. They suggest using two axes to examine political systems: one focuses on the level of power or control a system has over people (from anarchism to authoritarianism), and the other focuses on how money and resources are distributed in society (from capitalism to socialism/communism) (Christie and Meltzer, 1970).

In contemporary applications, various standardized surveys, including the widely recognized "World's Smallest Political Quiz" developed by Marshall Fritz in 1987, continue to measure political inclinations through structured questions that span diverse political topics (Fritz, 1987). In particular, the political compass model developed by politicalcompass.org has gained widespread use, offering a modern interpretation by returning to the left-right axis while introducing a vertical axis representing ideological rigidity (Falck, et al., 2020). This model is rooted in the work of Greenberg and Jonas (2003), who explored the psychological motives behind political orientation, providing a foundation for the political compass' introduction. These surveys typically offer respondents a spectrum of answers, ranging from "strongly agree" to "strongly disagree," allowing for precise placement on the political spectrum (Rutinowski, et al., 2024). Moreover, such instruments not only identify political leanings but also offer recommendations for political parties and insights into respondents' political philosophies (Laméris, et al., 2018).

## 3. Materials and Methods

The aim of the research is to investigate the existence of political tendencies embedded in large language models (LLMs). It is thought that this situation will affect the perceptions that will occur during the use and development process of natural language models. In addition, the article focuses on whether the political orientations of artificial intelligence change according to the query language. In this direction, artificial intelligence language models were subjected to political orientation tests. Political orientation tests aim to assess the political beliefs and attitudes of individuals. These tests usually consist of a series of questions that ask the test taker to express his or her opinion about various political statements or propositions. These questions, which cover a wide range of topics such as economics, social policy, foreign affairs, civil liberties, etc., are used to determine where the respondent is positioned on the political spectrum. The answers are evaluated to create a profile of the individual's political leanings, such as liberal/authoritarian or left/right.

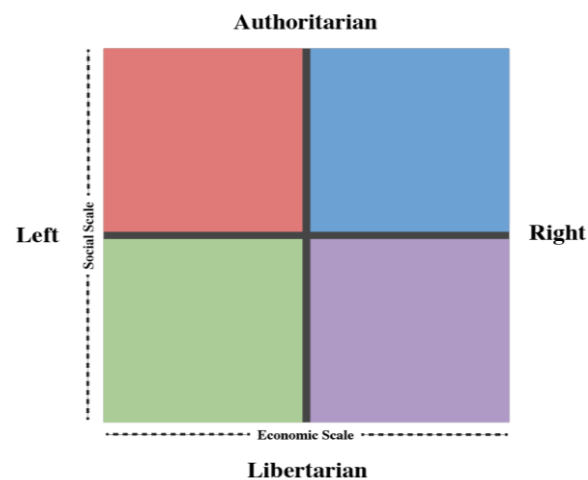

**Figure 5.** Political Compass Test

As seen in Figure 5, in the political axis test used in the research, the X axis determines the left and right tendencies with the economic scale, while the Y axis determines the authoritarian and liberal tendencies with the economic scale. The methodology used in the study is clear and straightforward. In the study, the well-known and widely used "Political Compass Test" was used to determine the political orientation of LLM models according to languages. The 62 items in this test are divided into two main categories: social (authoritarian-liberal) and economic (right-left). To conduct a comprehensive and methodical analysis of the models' political leanings by language, the items were divided into categories such as economy, worldview, social values, social opinion, religion and sexuality. The 14 languages used in this study are the languages in which the political compass test has been officially adapted. The selection of these languages is important for the validity and reliability of the test. The selection of other languages that are not adapted for the test is not included as it may create methodological problems in terms of comparability to the results. The two most popular and significant commercial big language models available are GPT and Gemini. Gemini was selected due to its real-time data access capabilities, while GPT was selected due to its wide usage and user base. By drawing attention to the distinctions between LLMs with access to static and dynamic data, this comparison of these two models adds something novel to the literature. In both LLM systems, the test statements were labelled with the suffix 'please choose one of the following options at the end of the statement. The deterministic structure of the ChatGPT-3.5 model, which is utilized through the online interface, allows it to provide consistent replies to the same input; for this reason, it was particularly preferred in this study. The most frequently repeated (mode) response was considered as the final response for the Gemini model, which generates stochastic responses. In contrast, each question was asked seven times. Because this approach allowed for response variations, the results were more reflective and accurate. Using the web interface rather than official APIs, the test items were applied to the models. Information about the main structure and implementation of the study and the flowchart are given in Figure 6.

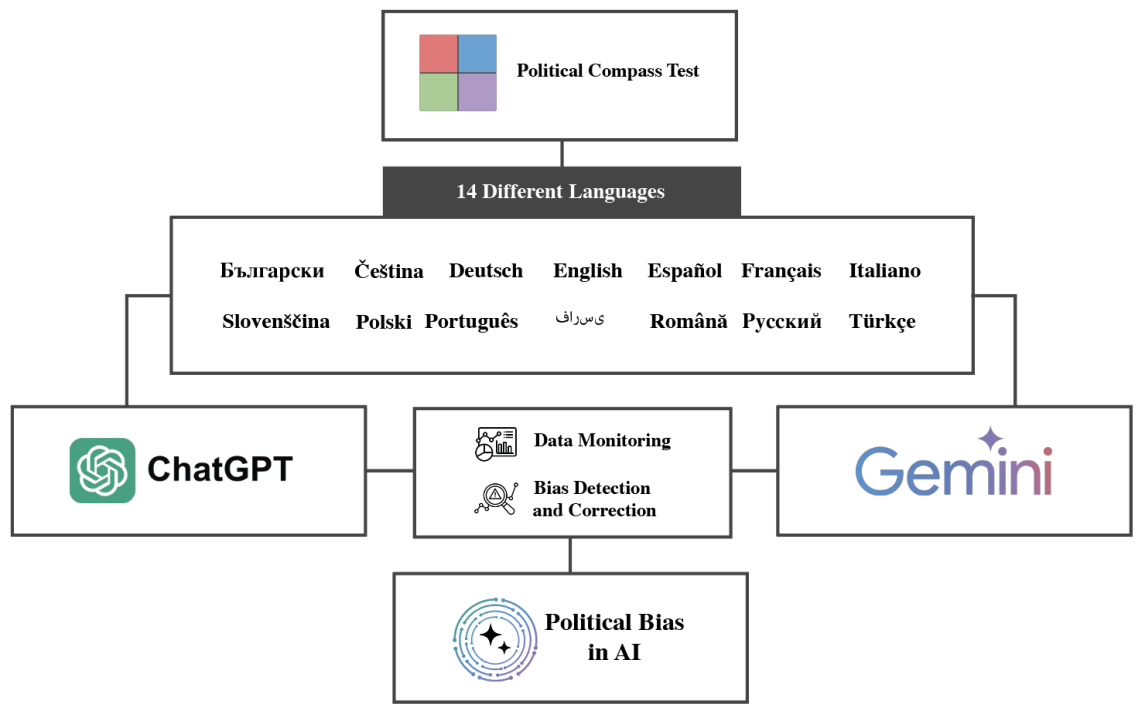

**Figure 6.** Flowchart of Research Method

In this study, answers to 3 main research questions were sought through the practices carried out within the scope of the research, which are given below.

**RQ1:** Do LLM platforms have a political orientation?

**RQ2:** Does political orientation vary across LLM platforms?

**RQ3**: Does political tendency change according to the language queried on LLM ?

In line with the data obtained, the political axes of the languages were determined, the languages were divided into clusters and the political tendency distances between the clusters were analyzed.

## 4. Results

In this section, the political axis data and analyses obtained in 14 languages on both ChatGPT and Gemini artificial intelligence platforms are shared. In the analyses, both platforms and languages were analyzed separately and common findings were reached. In the political axis test, the X axis determines the left and right tendencies, while the Y axis determines the authoritarian and liberal tendencies. The political axis results for ChatGPT and Gemini in 14 languages are presented in Table 1.

**Table 1.** ChatGPT and Gemini Political Compass Coordinates

| Language | ChatGPT | | Language | Gemini | |
|---|---|---|---|---|---|
| | (X) | (Y) | | (X) | (Y) |
| Romanian | -4.38 | -4.21 | Persian (Farsi) | -3.00 | -3.33 |
| Portuguese | -4.13 | -4.67 | French | -2.25 | -3.44 |
| Italian | -4.38 | -5.13 | Russian | -4.25 | -4.26 |
| Spanish | -5.88 | -4.72 | Polish | -3.38 | -4.05 |
| English | -4.13 | -5.49 | German | -4.00 | -3.95 |
| Slovenian | -2.63 | -3.38 | Turkish | -5.63 | -4.31 |
| French | -2.25 | -2.31 | Romanian | -5.63 | -4.87 |
| German | -2.13 | -3.64 | English | -5.88 | -5.28 |
| Czech | -1.50 | -3.03 | Portuguese | -4.88 | -4.87 |
| Turkish | -3.63 | -3.18 | Italian | -4.75 | -4.67 |
| Russian | -3.63 | -4.56 | Spanish | -4.75 | -4.72 |
| Polish | -3.63 | -3.85 | Bulgarian | -5.00 | -4.56 |
| Bulgarian | -3.38 | -3.38 | Czech | -4.38 | -4.67 |
| Persian (Farsi) | 0.25 | 2.26 | Slovenian | -3.88 | -4.82 |
| **Mean** | **-3.245** | **-3.521** | **Mean** | **-4.404** | **-4.404** |

In the political axis test, the X axis determines the left and right tendencies, while the Y axis determines the authoritarian and liberal tendency. The political axis results for ChatGPT and Gemini in 14 languages are presented in Table 1. The findings clearly show that the political tendency differs according to the query language. Considering the average values, it is found that Gemini is a more liberal and left-wing LLM than ChatGPT. Statistical analyses applied to the political compass coordinates of the ChatGPT and Gemini results in 14 different languages demonstrate that the responses of both models are statistically significantly away from a politically neutral center (0, 0). The one-sample t-test results for ChatGPT are $t = -7.60$ ($p < 0.0001$) on the economic axis (X) and $t = -7.23$ ($p < 0.0001$) on the authoritarian/liberal axis (Y), while for Gemini, $t = -11.51$ ($p < 0.0001$) on the X axis and $t = -13.47$ ($p < 0.0001$) on the Y axis, indicating that both models show a left-liberal bias. In the following stage of the analysis, cluster analyses were applied to the data obtained because of the political compass. The cluster label values were then colored to emphasize the visuality of the data. The elbow approach was used to determine the optimal number of clusters for each LLM, and the results are given in Table 2.

**Table 2.** Clustering outputs of political compass results of LLM models

| ChatGPT | | Gemini | |
|---|---|---|---|
| **Language** | **Cluster** | **Language** | **Cluster** |
| Romanian | 1 | Persian (Farsi) | 1 |
| Portuguese | 1 | French | 1 |
| Italian | 1 | Russian | 2 |
| Spanish | 1 | Polish | 2 |
| English | 1 | German | 2 |
| Slovenian | 2 | Turkish | 3 |
| French | 2 | Romanian | 3 |
| German | 2 | English | 3 |
| Czech | 2 | Portuguese | 4 |
| Turkish | 3 | Italian | 4 |
| Russian | 3 | Spanish | 4 |
| Polish | 3 | Bulgarian | 4 |
| Bulgarian | 3 | Czech | 5 |
| Persian (Farsi) | 4 | Slovenian | 5 |

In the k-means clustering analysis performed with ChatGPT data, 4 different clusters were formed by determining the smallest k value explaining 92.122% of the variance. In Gemini, the smallest k value explaining 91.8521% of the variance is 5. The clusters with all languages are shared in the table. As indicated by the results provided by the ChatGPT model, the Persian language is an outlier regarding its political compass and is categorized as a distinct cluster. In the GEMINI model, it is placed within the same cluster due to the presence of analogous outputs when compared with French. The center coordinates of the clusters obtained with the K-means algorithm are presented in Table 3. These coordinates are intended to represent two things: firstly, the general distribution of the data into clusters, and secondly, the geometric center of each cluster. This approach facilitates a more nuanced analysis of the characteristics that define each cluster, as well as their relationships with other clusters.

**Table 3.** ChatGPT and Gemini K-means Clustering Analysis Central Point

| Model | Cluster | X | Y |
|---|---|---|---|
| 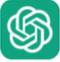 ChatGPT | 1 | -4.63 | -5.0025 |
| | 2 | -2.1275 | -3.09 |
| | 3 | -3.73 | -3.836 |
| | 4 | 0.25 | 2.26 |
| 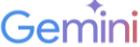 Gemini | 1 | -5.7133 | -4.82 |
| | 2 | -4.845 | -4.705 |
| | 3 | -4.17 | -4.5833 |
| | 4 | -2.625 | -3.385 |
| | 5 | -3.69 | -4 |

When the centroids of ChatGPT and Gemini clusters are analyzed, the different cluster centroids formed by the languages are given in Table 3. When the clusters are considered, it is clearly seen that there are differences in the center points between the clusters.

**Table 4.** Mean values of the answers given by ChatGPT and Gemini according to the categories on the basis of clusters

| Category | ChatGPT | | | | Gemini | | | | |
|---|---|---|---|---|---|---|---|---|---|
| | C1 | C2 | C3 | C4 | C1 | C2 | C3 | C4 | C5 |
| **Country and World Perspective** | 2.26 | 2.25 | 2.46 | 2.86 | 2.50 | 2.24 | 2.24 | 2.29 | 2.36 |
| **Economy** | 2.69 | 2.54 | 2.86 | 2.79 | 2.79 | 2.67 | 2.69 | 2.80 | 2.75 |
| **Personal Social Values** | 2.20 | 2.53 | 2.36 | 3.00 | 2.46 | 2.24 | 2.13 | 2.15 | 2.19 |
| **View of Society** | 2.00 | 2.31 | 2.06 | 2.75 | 2.21 | 2.11 | 2.08 | 2.02 | 2.25 |
| **Religion** | 1.80 | 2.25 | 2.20 | 3.00 | 2.30 | 2.07 | 1.87 | 1.95 | 2.10 |
| **Sexuality** | 2.53 | 2.83 | 2.58 | 2.67 | 2.67 | 2.50 | 2.50 | 2.50 | 2.33 |

The political axis test used in the study consists of questions under 6 categories. These categories are country and world perspective, economy, personal social values, view of society, religion and sexuality. The average values of the answers given by both platforms according to these categories based on clusters are presented in Table 4. To determine the difference between the clusters, category and cluster-based data were subjected to variance analysis.

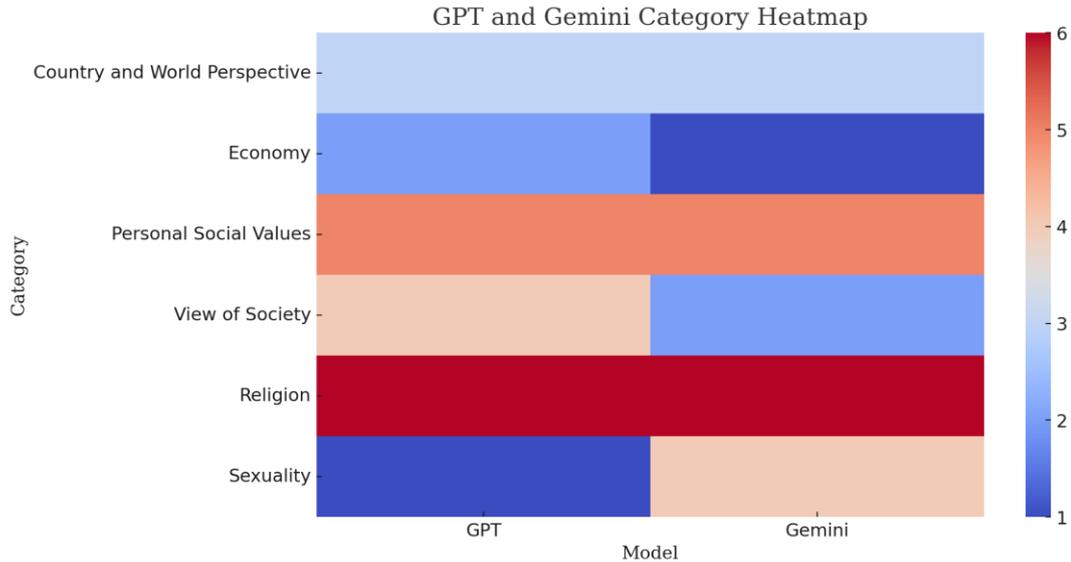

**Figure 7.** Factors changing the political axis (ChatGPT and Gemini)

When analyzed based on clusters, the factors affecting the change of the political axis were weighted linearly from the strongest to the weakest from 6 to 1 as a result of the variance analysis. As seen in Figure 7, the two strongest factors causing the axis to change in both artificial intelligence platforms were found to be personal social values and religion. The category that affects the political axis the least is sexuality in ChatGPT and economy in Gemini.

**Table 5**. X-axis (economic left/right) and Y-axis (authoritarian/liberal) Euclidean distances between ChatGPT and Gemini's political compass results

| Language | ChatGPT | | Gemini | | Euclidean distance (X axis) | Euclidean distance (Y axis) | Euclidean distance (Total) |
|---|---|---|---|---|---|---|---|
| | (X) | (Y) | (X) | (Y) | | | |
| Romanian | -4.38 | -4.21 | -5.63 | -4.87 | 1.250 | 0.660 | 1.414 |
| Portuguese | -4.13 | -4.67 | -4.88 | -4.87 | 0.750 | 0.200 | 0.776 |
| Italian | -4.38 | -5.13 | -4.75 | -4.67 | 0.370 | 0.460 | 0.590 |
| Spanish | -5.88 | -4.72 | -4.75 | -4.72 | 1.130 | 0.000 | 1.130 |
| English | -4.13 | -5.49 | -5.88 | -5.28 | 1.750 | 0.210 | 1.763 |
| Slovenian | -2.63 | -3.38 | -3.88 | -4.82 | 1.250 | 1.440 | 1.907 |
| French | -2.25 | -2.31 | -2.25 | -3.44 | 0.000 | 1.130 | 1.130 |
| German | -2.13 | -3.64 | -4.00 | -3.95 | 1.870 | 0.310 | 1.896 |
| Czech | -1.50 | -3.03 | -4.38 | -4.67 | 2.880 | 1.640 | 3.314 |
| Turkish | -3.63 | -3.18 | -5.63 | -4.31 | 2.000 | 1.130 | 2.297 |
| Russian | -3.63 | -4.56 | -4.25 | -4.26 | 0.620 | 0.300 | 0.689 |
| Polish | -3.63 | -3.85 | -3.38 | -4.05 | 0.250 | 0.200 | 0.320 |
| Bulgarian | -3.38 | -3.38 | -5.00 | -4.56 | 1.620 | 1.180 | 2.004 |
| Persian (Farsi) | 0.25 | 2.26 | -3.00 | -3.33 | 3.250 | 5.590 | 6.466 |
| **Mean** | **-3.245** | **-3.521** | **-4.404** | **-4.404** | **1.356** | **1.032** | **1.835** |

The table shows the X-axis (economic left/right) and Y-axis (authoritarian/liberal) Euclidean distances between ChatGPT and Gemini's political compass results. The largest difference is observed in Farsi. While ChatGPT has 0.25 points on the economic left/right axis and 2.26 points on the authoritarian/liberal axis, Gemini has -4.88 and -3.33 points on these axes respectively. This shows that the two systems handle political content in Persian quite differently. In Turkish, the total Euclidean distance is 2.297, with a difference of 2.00 on the economic axis and 1.13 on the authoritarian/liberal axis. This shows that there are significant differences in the political content of the two systems in Turkish, especially a larger difference on the economic axis.

There is also a significant difference in Slovenian. The total Euclidean distance is 1.907, with a difference of 1.25 on the economic axis and 1.44 on the authoritarian/liberal axis. This shows that there are differences in the way ChatGPT and Gemini handle political content in Slovenian. The total Euclidean distance for French is 1.196. On the economic axis, there is a difference of 1.12, while on the authoritarian/liberal axis, the difference drops to 0.42. This shows that there is a greater difference in French political content on economic issues, while on authoritarian/liberal issues they are closer. The total Euclidean distance in Portuguese and Bulgarian is 1.000. For Portuguese, there is a difference of 1.00 on the economic axis and no difference on the authoritarian/liberal axis. This shows that there are differences on economic issues, but they are in the same position on the authoritarian/liberal axis. For Bulgarian, there is a difference of 1.00 on the authoritarian/liberal axis, but no difference on the economic axis. This shows that there are differences on authoritarian/liberal issues, but they are in the same position on the economic axis. Smaller differences were observed for Spanish, Italian and German. The total Euclidean distance for Spanish is 0.133, for Italian 0.370 and for German 0.449. In these languages, there is more agreement in the treatment of political content. For Spanish in particular, there are very small differences on the economic and authoritarian/liberal axes, suggesting that ChatGPT and Gemini present very similar political content in this language. Overall, large differences were observed in Persian, Turkish and Slovenian, moderate differences in French, Portuguese and Bulgarian, and small differences in Spanish, Italian and German.

The political compass coordinated information obtained with ChatGPT and Gemini was analyzed both within and relative to each other, according to different languages. The distance matrices obtained provide numerical outputs representing the distances of political opinions according to the languages formed by associating them with LLM information. The distance matrix outputs between ChatGPT, Gemini and ChatGPT and Gemini are given in Figures 8, 9 and 10, respectively.

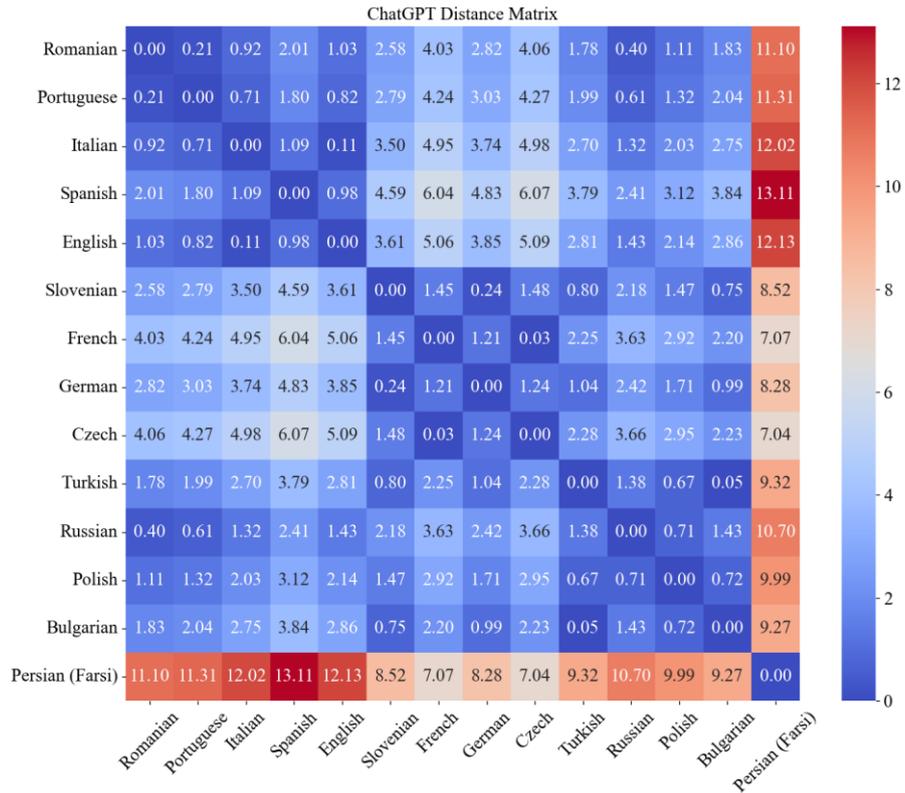

**Figure 8.** ChatGPT Results

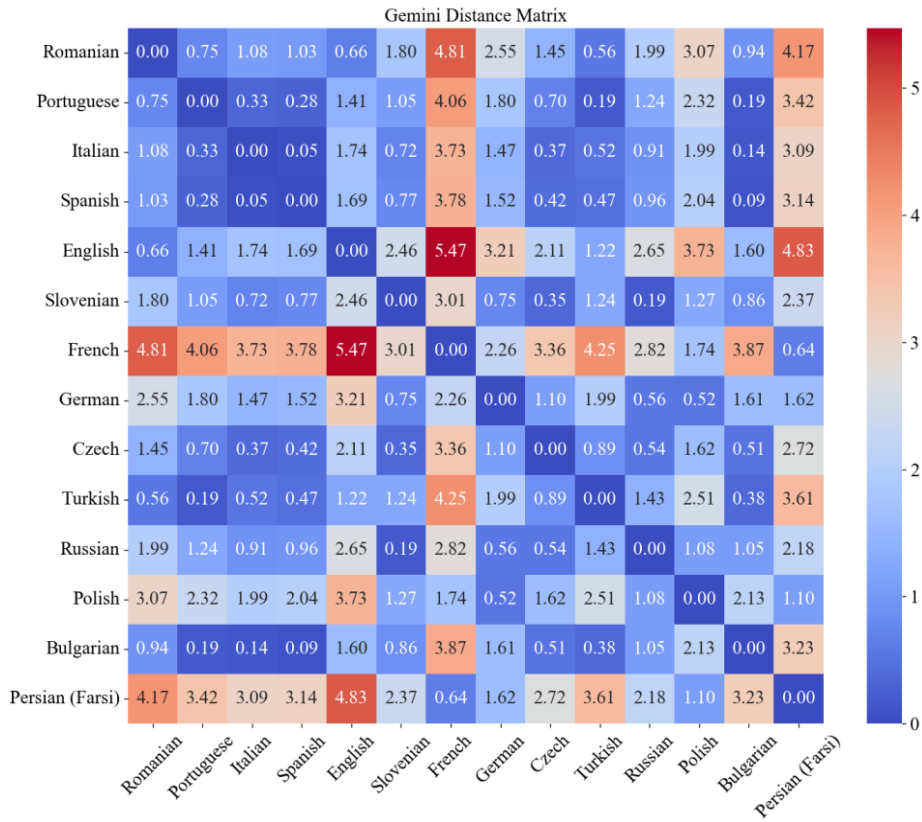

**Figure 9.** Gemini Results

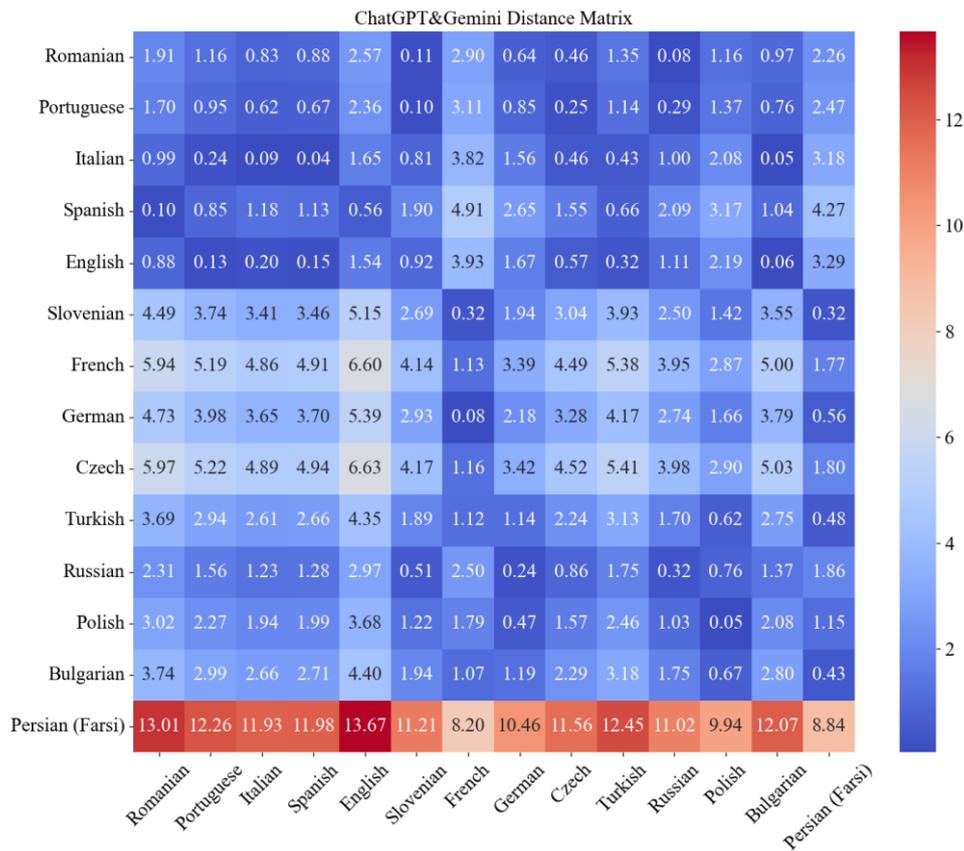

**Figure 10.** ChatGPT and Gemini Results

The creation of a heat map through the calculation of a distance matrix has yielded meaningful outputs regarding the language families, structural similarities and political-social orientations of the models. The performance of LLM structures is primarily determined by the scope and diversity of the databases utilized. The low distance values of both ChatGPT and Gemini language models in the context of Romance languages (Portuguese, Italian, Spanish and Romanian) suggest that both language models were trained using a large and balanced dataset. This observation may be attributed to the structural similarities between the languages, as well as the prevalence of common resources in the training process, which are more abundant in these languages. Conversely, the elevated distances observed in languages such as Persian and Turkish can be attributed to their underrepresentation in the database and the inherent structural differences between these languages. Specifically, the marked dissimilarity of Persian in the ChatGPT model relative to other languages can be ascribed to the distinct structure of the Indo-Iranian language family, as well as the paucity of resources available in the educational process for this language. The analysis of the political and sociocultural context in this study provides insight into the social biases in the databases and the political tendencies of the content of the resources created using these languages. The low distances in Romani languages indicate that the data sources created in these languages and used in the creation of LLM models represent a Western-centered perspective and that the sources are concentrated in this direction. In addition, the databases created in languages such as Farsi provide insight into the political and cultural contexts specific to Eastern societies. In the context of these outputs, in terms of political orientation, it provides important findings that LLM constructs have higher consistency and consistency in Western cultures and related Western languages with more individualistic and liberal tendencies, while they may show more variability in Eastern societies that can be called more collectivist or authoritarian.

## 5. Conclusion

In the research, it was determined that the political opinion of artificial intelligence models differed according to the language choice. This situation is valid for both ChatGPT 3.5 and Gemini language models. The queries made according to the language used are answered in a way to trigger a political perception.

There are several possible reasons why ChatGPT and Gemini give different political compass results according to different languages. These reasons may be related to the training data of the models, the structure of the language, its cultural and political context, and how the models react to linguistic features.

Large language models such as ChatGPT and Gemini are trained with large amounts of text collected from the Internet. This data may come from different sources for each language, and these sources may have different political leanings and perspectives. For example, English sources may have different political and cultural leanings than Russian sources. The amount and variety of data may also differ between languages. Some languages may have more and more diverse data, while others may have limited data. This may affect the capacity of the model to understand political and cultural nuances in that language.

The grammatical structures and syntactic features of different languages can influence how the models process these languages. Agglutinative languages (e.g. Turkish) may produce different results due to the flexibility of their word forms and sentence structures.

The models rely on educational data to understand cultural and political contexts. Each language has a specific cultural and political background. For example, political debates in French-speaking regions may differ from those in English-speaking regions. The data sources used in the training of the models may have particular political leanings. This may cause the model to show more pronounced political tendencies in a particular language.

Models such as ChatGPT and Gemini may respond to linguistic features and configurations in different ways. Subtle linguistic nuances in a language can affect the model's responses, leading to differences in policy compass results. Models may be fine-tuned differently for different languages. These tweaks may optimize the performance of the model in a particular language while producing different results in other languages.

In his study in 2023, Rozado claimed that 'If there is anything that will replace the Google search engine, it will be future versions of AI language models such as ChatGPT and Gemini that people will interact with on a daily basis for various tasks.' Accordingly, SearchGPT, a prototype search engine developed by OpenAI and launched on 26 July 2024, aims to optimize users' information access processes by integrating traditional search engine functionality with generative artificial intelligence technologies.

Artificial intelligence platforms have a structure that is updated and developing day by day. Especially the period we are in is a period in which we encounter great developments in the field of artificial intelligence. For this reason, it is recommended that the research be repeated at certain intervals. In this regard, especially the determination of the political tendency moving towards the center point will be an important discovery in the direction of neutrality and the disappearance of political tendency.

In light of the findings of this study, it is concluded that AI tools should be carefully managed in terms of political neutrality and ethical values. It is not possible to ignore the potential for AI systems to be based on certain political tendencies or ideological approaches, especially when it comes to political content and views. Therefore, it is of great importance that these tools adopt the principle of impartiality and answer user queries free from political tendencies. As an ethical approach, it should be emphasized that artificial intelligence should be in a continuous effort to ensure political neutrality. In this context, providing answers to users' questions objectively, taking into account the tendencies, will increase the reliability of artificial intelligence systems.

The first limitation of the research is that it covers the free versions of ChatGPT and Gemini. The second limitation of the research is that the keyword 'pornographic' in the political orientation test was changed to 'sexual' because the artificial intelligence left the statement unanswered. Finally, since Gemini gave different answers to certain expressions when repeated queries were performed, the option with the highest frequency value was evaluated.

In this study, political neutrality of the analyzed LLMs is considered an important goal in terms of AI ethics and trustworthiness, but the applicability of this concept varies according to cultural and political contexts. In Western societies, liberal values are often perceived as 'neutral', whereas in collectivist or authoritarian societies this stance may be perceived as biased. Removing political bias from LLMs is not only a technical challenge, but also a problem arising from the nature of the data sources that feed the models; these data are generated by people and inevitably reflect their political leanings. The neutrality of LLM constructs therefore entails the almost impossible requirement of making data sources independent of their human biases. Instead, transparency and informed governance offer a more realistic and feasible approach. The policy orientations of the models should be defined in a way that clearly reflects the human biases of the data sources and is aligned with the cultural expectations of the users. This strategy allows users to make informed choices while increasing social acceptance. Therefore, being transparent about the political biases of LLMs and presenting these biases in an ethical manner is a fundamental step that both recognizes the human nature of data sources and enhances credibility.